\documentclass[10pt,twocolumn,letterpaper]{article}

\usepackage{iccv}
\usepackage{times}
\usepackage{epsfig}
\usepackage{graphicx}
\usepackage{amsmath}
\usepackage{amssymb}

\usepackage{booktabs}
\usepackage{multirow}
\usepackage{makecell}
\usepackage{bbding}

\usepackage{algorithm}
\usepackage{algorithmic}

\usepackage[pagebackref=true,breaklinks=true,letterpaper=true,colorlinks,bookmarks=false]{hyperref}

\iccvfinalcopy 

\def\iccvPaperID{1123} 

\begin{document}

\title{Learning Structure-Guided Diffusion Model for 2D Human Pose Estimation}

\author{Zhongwei Qiu$^{1,3}$
\and
Qiansheng Yang$^2$
\and
Jian Wang$^2$
\and
Xiyu Wang$^3$
\and
Chang Xu$^3$
\and
Dongmei Fu$^1$
\and
Kun Yao$^2$
\and
Junyu Han$^2$
\and
Errui Ding$^2$
\and
Jingdong Wang$^2$
\and
$^1$ University of Science and Technology Beijing, $^2$ Baidu, 
$^3$ University of Sydney
}

\maketitle
\ificcvfinal\thispagestyle{empty}\fi

\begin{abstract}
One of the mainstream schemes for 2D human pose estimation (HPE) is learning keypoints heatmaps by a neural network.
Existing methods typically improve the quality of heatmaps by customized architectures, such as high-resolution representation and vision Transformers.
In this paper, we propose \textbf{DiffusionPose}, a new scheme that formulates 2D HPE as a keypoints heatmaps generation problem from noised heatmaps.
During training, the keypoints are diffused to random distribution by adding noises and the diffusion model learns to recover ground-truth heatmaps from noised heatmaps with respect to conditions constructed by image feature.
During inference, the diffusion model generates heatmaps from initialized heatmaps in a progressive denoising way.
Moreover, we further explore improving the performance of DiffusionPose with conditions from human structural information.
Extensive experiments show the prowess of our DiffusionPose, with improvements of 1.6, 1.2, and 1.2 mAP on widely-used COCO, CrowdPose, and AI Challenge datasets, respectively.

\end{abstract}


\section{Introduction}
Human pose estimation is a fundamental computer vision task, which requires localizing the keypoints coordinates from the human body in an image and has many applications in action recognition~\cite{liu2020disentangling}, human reconstruction~\cite{zheng2019deephuman}, and human neural radiance fields~\cite{xu2021h}, etc. The modern approaches for 2D human pose estimation can be categorized as top-down~\cite{xiao2018simple,sun2019deep,xu2022vitpose}, bottom-up~\cite{kreiss2019pifpaf,cheng2020higherhrnet,geng2021bottom}, and one-stage methods~\cite{zhou2019objects,shi2022end,yang2023explicit}. 
Usually, the top-down methods show better performances than other paradigms due to the unified scale of persons and higher resolutions of people. And we focus on top-down manner in this paper.

\begin{figure}[t]
  \centering
  \includegraphics[width=\linewidth]{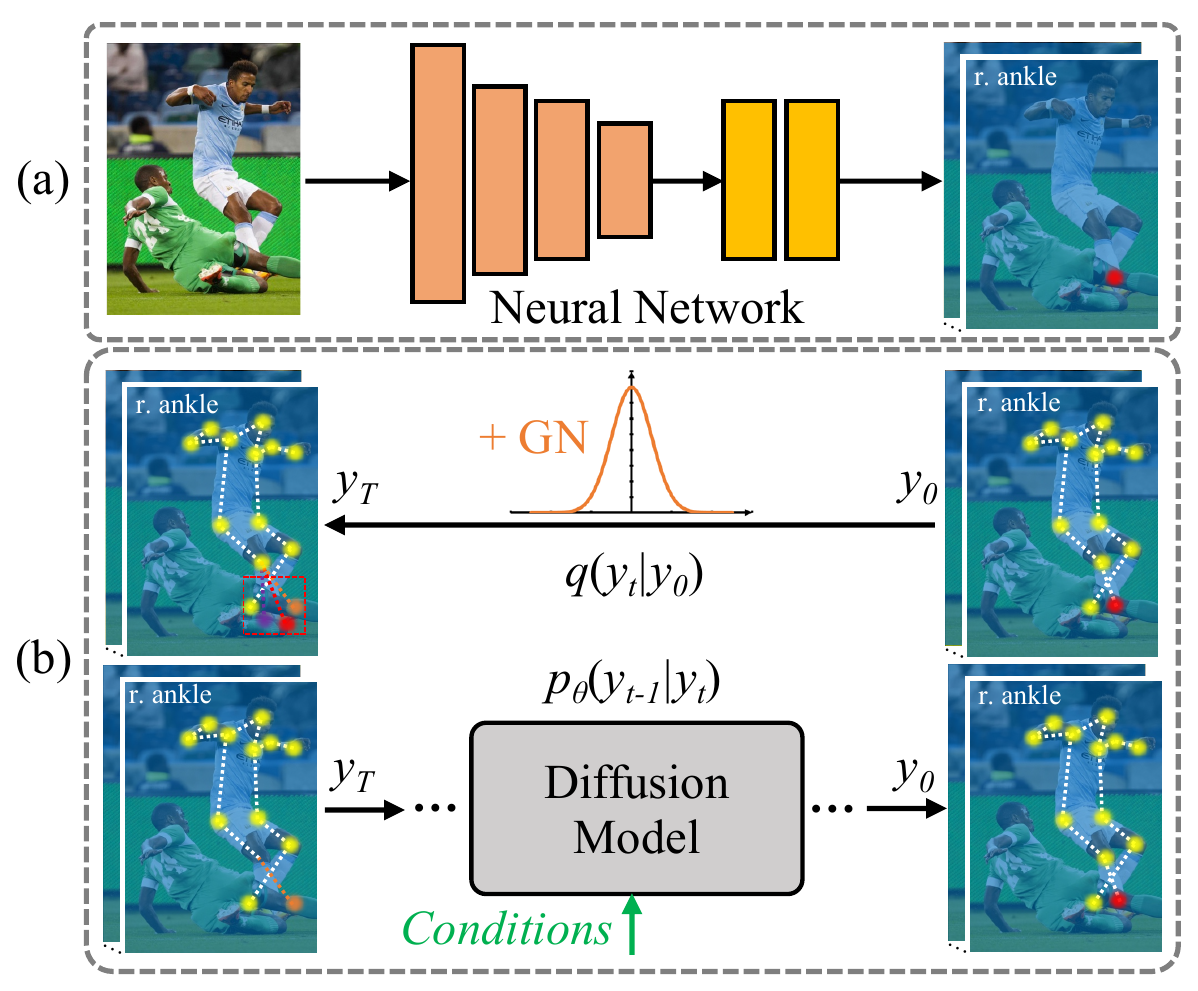}

  \caption{(a) Existing heatmap-based pose estimation framework~\cite{xiao2018simple}. (b) Diffusion-generation scheme, which includes a forward diffusion process $q$ by adding Gaussian Noises (GN) and the reverse process $p_\theta$ of recovering heatmaps from noised heatmaps.}
  \label{fig:teaser}
\end{figure}

Learning heatmaps is widely adopted in top-down~\cite{xiao2018simple,sun2019deep,xu2022vitpose} and bottom-up methods~\cite{kreiss2019pifpaf,cheng2020higherhrnet} because it can reserve the spatial information and make the model easy to converge.
To obtain better heatmap representation, existing methods typically design customized architectures to extract better features, such as high-resolution representation~\cite{sun2019deep,cheng2020higherhrnet} and vision Transformers~\cite{yang2021transpose,li2021tokenpose,xu2022vitpose}.
For example, HRNet~\cite{sun2019deep} maintains high-resolution features during each downsampling stage of the neural network. ViTPose~\cite{xu2022vitpose} utilizes a full vision Transformer network to extract stronger feature representation.
As shown in Figure~\ref{fig:teaser} (a), these methods all follow a discriminative scheme of detecting each pixel of an image.
Besides, some refinement-based methods~\cite{huang2020devil,zhang2020distribution,moon2019posefix} improve the accuracy of coordinates outputted from heatmaps by a two-stage refining model~\cite{moon2019posefix}, post-process
based on heatmap distribution~\cite{zhang2020distribution}, and learning keypoints offsets
relative to heatmap peaks~\cite{huang2020devil}.
However, the performances of these post-process refinements
are inevitably impacted by the quality of the original heatmaps. 
Our motivation is to explore a novel scheme for generating high-quality heatmaps.

Recently, diffusion models~\cite{ho2020denoising,song2020denoising} have shown superior performance in multi-modal content generation tasks, such as text-guided image/video generation~\cite{nichol2022glide,ho2022video}, image super-resolution~\cite{li2022srdiff,saharia2022image}, 3D shapes and textures synthesis~\cite{metzer2022latent}, and so on.
These successes are from the strong ability of the diffusion model that recovers the signals from noised inputs in a progressive way. Inspired by that, researchers successfully apply the diffusion denoising framework to object detection~\cite{chen2022diffusiondet} and segmentation~\cite{gu2022diffusioninst,brempong2022denoising,baranchuk2021label}, which brings a promising prospect of tackling visual perception tasks via diffusion models. However, few explorations have been studied on applying diffusion models to 2D human pose estimation.

To generate high-quality heatmaps for 2D human pose estimation, we propose DiffusionPose, which formulates the 2D human pose estimation as a generative process of heatmaps from noised heatmaps via diffusion model.
The novel diffusion-generation scheme for 2D human pose estimation is shown in Figure~\ref{fig:teaser} (b), which includes a forward diffusion process $q(y_t|y_0)$ and a reverse process $p_\theta(y_{t-1}|y_{t})$.
During training, the keypoints are diffused to random distribution by adding Gaussian noises and the diffusion model learns to recover ground-truth heatmaps from noised heatmaps with respect to conditions constructed by image feature.
During inference, the diffusion model generates high-quality heatmaps from initialized heatmaps in a progressive denoising way, with the guidance of conditions.
Furthermore, we explore improving the performance of DiffusionPose by incorporating various human structural information as conditions and studying the influences of different heatmaps resolutions on DiffusionPose.
These components lead DiffusionPose to achieve favorable results on three 2D pose estimation datasets.

Our main contributions can be summarized as follows:
\begin{itemize}
    \item We propose DiffusionPose, which formulates the 2D human pose estimation as a denoising process from noised heatmaps via diffusion model. To the best of our knowledge, this is the first successful application of the diffusion model in 2D human pose estimation.
    \item We further propose the structure-guided diffusion decoder (SGDD) and high-resolution SGDD to reform the diffusion model for human pose estimation, which leads DiffusionPose to achieve better performance.
    \item Extensive experiments on COCO, CrowdPose, and AI Challenge datasets show that DiffusionPose achieves favorable results against the previous schemes.
\end{itemize}

\section{Related Work}
\subsection{Human Pose Estimation}
\textbf{2D Human Pose Estimation.}
Modern 2D human pose estimation methods~\cite{li2021human,qiu2019learning,sun2019deep,xu2022vitpose} usually adopt neural networks to extract deep features, then estimate human poses from deep features. Therefore, related works mainly focus on the backbone network~\cite{xiao2018simple,qiu2020dgcn,wang2022multi}, pose representation~\cite{newell2016stacked,li2021human}, and the framework of multi-person pose estimation~\cite{shi2022end,qiu2023psvt}.
Hourglass~\cite{newell2016stacked} stacks a multi-stage CNN network to extract features and regress 2D keypoints heatmaps to localize the coordinates of human joints. Based on the heatmap representation, SimpleBaseline~\cite{xiao2018simple}, HRNet~\cite{sun2019deep}, and ViTPose~\cite{xu2022vitpose} propose different backbone networks to extract better features, respectively. Except for the heatmap-based representation, some regression-based methods~\cite{sun2017compositional,li2021human} also explore the different expressions of 2D human poses. To tackle the multiple instances in an image, three different frameworks are proposed. Top-down methods~\cite{xiao2018simple,sun2019deep,xu2022vitpose} first detect human bounding boxes, then apply the single-person pose models on the cropped images with human boxes. Bottom-up methods~\cite{cao2017realtime,cheng2020higherhrnet} first estimate human poses, then assign them to different persons. One-stage methods~\cite{wei2020point,shi2022end} directly regress multi-person poses from an image by keypoints offsets or pose tokens in a single stage. 

\textbf{Pose Refinement.}
Some methods~\cite{fieraru2018learning,li2019crowdpose,wang2020graph,zhang2020distribution,wang2022contextual} refine learned human poses by structure or distribution information. Graph-PCNN~\cite{wang2020graph} adopts a two-stage graph pose refinement module to refine the human pose from sampled keypoints on heatmaps. PoseFix~\cite{moon2019posefix} proposes a model-agnostic general human pose refinement network to refine the estimated 2D pose from any pose model. DARK~\cite{zhang2020distribution} proposes a distribution-aware coordinate representation to modulate 2D heatmaps. Although great improvements have been achieved through these approaches, their successes still rely on the quality of original heatmaps generated by the first stage.
In this paper, we aim to propose a new scheme for generating heatmaps via diffusion models.

\subsection{Diffusion Model}
\textbf{Diffusion Model for Generation Tasks.}
Recently, diffusion models~\cite{ho2020denoising,song2020denoising,song2020score} have achieved superior performance on the generation tasks, such as image/video synthesis~\cite{nichol2022glide,ho2022video}, image super-resolution~\cite{li2022srdiff,saharia2022image,qiu2022learning}, 3D shape and synthesis~\cite{metzer2022latent}, etc. 
Diffusion models add noises to the signals and learn to recover signals from noised data samples by a progressive denoising process. Although diffusion achieves great success on many generation tasks, it has huge costs on computational resources. Thus, DDIM~\cite{song2020denoising} and its variant~\cite{wang2023learning} transform the denoising process to the non-Markov process from the Markov process by skipping-step sampling, which improves the inference speed and booms the application of diffusion models on other computer vision tasks.

\textbf{Diffusion Model for Perception Tasks.}
With the success of diffusion models on generation tasks, researchers pay attention to applying diffusion models to other perception tasks, such as object detection~\cite{chen2022diffusiondet}, segmentation~\cite{gu2022diffusioninst,brempong2022denoising,baranchuk2021label,chen2022generalist}, 3D pose lifting~\cite{gong2022diffpose,holmquist2022diffpose}, and so on. DiffusionDet~\cite{chen2022diffusiondet} diffuses ground-truth object bounding boxes to noised boxes, then utilizes the diffusion model to recover ground-truth bounding boxes from randomly sampled boxes. Based on DiffusionDet, DiffusionInst~\cite{gu2022diffusioninst} adds a segmentation head on the regression head of DiffusionDet, which achieves instance segmentation by the diffusion model. For 3D pose lifting, some works~\cite{gong2022diffpose,holmquist2022diffpose} start from a randomly sampled 3D pose and recover the 3D pose with the guidance of 2D pose context. The success of diffusion models on perception tasks inspires us to study how to apply the diffusion model in 2D pose estimation and utilize its strong ability of recovering signals from noises to generate high-quality keypoints heatmaps.

\section{Approach}
\subsection{Preliminary}
Diffusion models achieve great success on content generation tasks~\cite{ho2020denoising,ho2022video,saharia2022image}, which includes the forward diffusion process and reverse denoising process. Given a data distribution $y_0$, the forward process $q$ adds Gaussian noise to $y_0$ over $T$ steps:
\begin{equation}
    q(y_t|y_{t-1}) = \mathcal{N}(y_t|\sqrt{\alpha _t}y_{t-1}, (1-\alpha_t)\textit{\textbf{I}}),
\end{equation}
where $0<\alpha_t <1$ is the weight hyper-parameter, which controls the variance of the added noise. Particularly, given $y_0$, the sampled $y_t$ can be generated as:
\begin{equation}
\label{eq:q_sample}
    q(y_t|y_0) = \mathcal{N}(y_t| \sqrt{\gamma_t}y_0, (1-\gamma_t)\textit{\textbf{I}}),
\end{equation}
where $\gamma_t = \prod^t_{i=1}\alpha_i$. Practically, Equation \ref{eq:q_sample} can be recapped as:
\begin{equation}
\label{eq:add_noise}
    y_t = \sqrt{\gamma_t}y_0 + \sqrt{1-\gamma_t}\epsilon, \epsilon \thicksim \mathcal{N}(\textit{\textbf{0}}, \textit{\textbf{I}}).
\end{equation}
where $\epsilon$ represents the sampled noises.

Recently, many works~\cite{chen2022diffusiondet,gu2022diffusioninst,li2022srdiff} have applied the diffusion models to the perception tasks in computer vision.
For object detection~\cite{chen2022diffusiondet} and instance segmentation~\cite{gu2022diffusioninst}, a neural network $f_\theta(y_t, t, x)$ is trained to estimate $y_0$ from noised $y_t$ during the training process.
During the inference, the $y_0$ is estimated from noised $y_t$ by diffusion model $f_\theta$, conditioned on the corresponding image $x$.

In this work, we formulate the 2D human pose estimation as the task of 2D heatmap generation by the diffusion model. During training, the human keypoints $y_0$ are diffused to noised points $y_t$, which is further used to generate keypoints mask $y^{mask}_t$. Then, the feature condition $x^{c}$ is extracted from image $x$ based on $y^{mask}_t$. With the guidance of $x^{c}$, the diffusion model learns to recover ground-truth heatmaps $y_0^{hm}$ from noised heatmaps $y^{hm}_t$. 
During inference, starting from noised heatmaps $y^{hm}_t$, the diffusion model estimates $y_0^{hm}$ in a progressive way, which is eventually used to decode ground-truth keypoints $y_0$.

\begin{figure*}[t]

  \centering
  \includegraphics[width=0.99\linewidth]{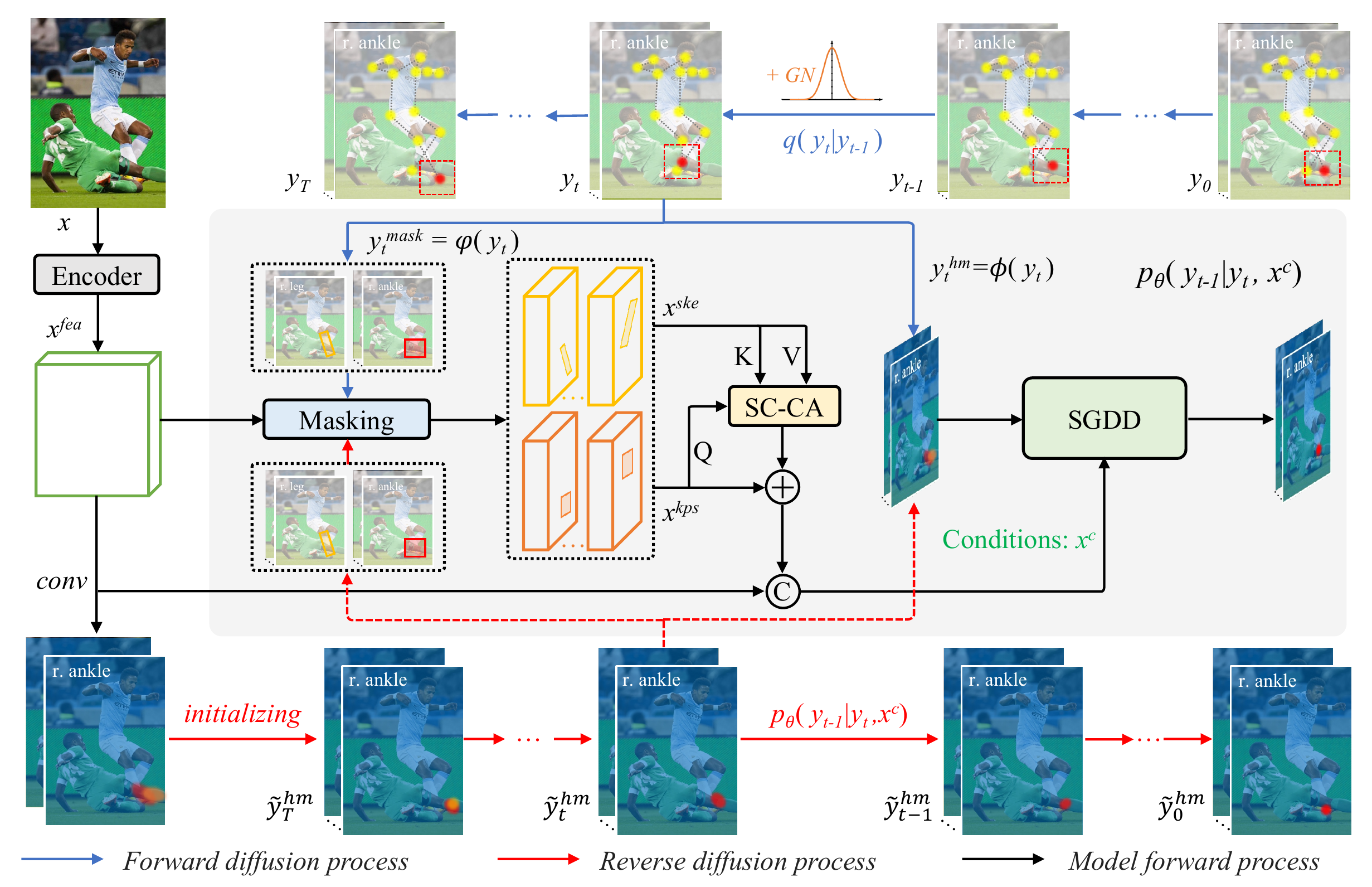}

  \caption{The framework of DiffusionPose, which includes the forward diffusion process (FDP), the model forward process (MFP), and the reverse diffusion process (RDP). During FDP, given step $t$, keypoints $y_0 \in \mathbb{R}^{J\times 2}$ diffuse to noised keypoints $y_t$ by $q(\cdot)$ with Gaussian Noises (GN). $y_t$ is used to generate mask $y_t^{mask}$ by operation $\varphi(\cdot)$, which includes keypoints masks $M^{kps}$ and skeleton masks $M^{ske}$. $y_t$ is also used to generate the noised heatmaps $y_t^{hm}$ by operation $\phi(\cdot)$. Based on $y_t^{mask}$, keypoints feature $x^{kps}$ and skeleton feature $x^{ske}$ are generated by masking $x^{fea}$ from encoder. $x^{kps}$ and $x^{ske}$ are sent into SC-CA module to generate feature conditions $x^c$. The diffusion decoder (SGDD) learns to recover keypoints heatmaps from noised heatmaps. During RDP, SGDD estimates heatmaps from noised heatmaps in a progressive denoising way as $p_{\theta}(y_{t-1}|y_t, x^c)$. $\oplus$ and circled $C$ are summing and concatenating operations, respectively.}
\label{fig:framework}
\end{figure*}

\subsection{DiffusionPose}
The framework of DiffusionPose is shown in Figure~\ref{fig:framework}, which contains the forward diffusion process (FDP), the model forward process (MFP), and the reverse diffusion process (RDP).
As shown in Figure~\ref{fig:framework}, 
the blue flow lines, black flow lines, and red flow lines indicate the FDP, MFP, and RDP, respectively.
During training, FDP diffuses ground-truth keypoints to noised keypoints by adding Gaussian noises, further generating noised heatmaps and feature masks.
Then, MFP extracted features from the input image with feature masks as the conditions for the diffusion model. The decoder of DiffusionPose learns to recover ground-truth heatmaps from noised heatmaps with the guidance of conditions.
During inference, RDP recovers the heatmaps from initialized heatmaps by diffusion decoder in a progressive denoising way and decodes the keypoints coordinates from estimated heatmaps.

\subsubsection{Forward Diffusion Process}

The Forward Diffusion Process (FDP) is shown in Figure \ref{fig:framework} (Blue flow process) and Algorithm \ref{alg:train}.
Given ground-truth keypoints $y_0\in \mathbb{R}^{J\times 2}$ and time step $t\in [1,T]$, FDP generates noised heatmaps $y_t^{hm} \in \mathbb{R}^{J\times H\times W}$ and feature masks $y_t^{mask}$. The diffusion process can be formulated as:
\begin{equation}
\label{eq:q_sample_training}
\begin{aligned}
    y_t &= q(y_t|y_0, \zeta) \\
&=\sqrt{\gamma_t}(\zeta \cdot y_0) + \sqrt{1-\gamma_t}\epsilon, \epsilon \thicksim \mathcal{N}(\textit{\textbf{0}}, \textit{\textbf{I}}), 
\end{aligned}
\end{equation}
where $q(\cdot)$ is the process of sampling in Equation \ref{eq:q_sample} and \ref{eq:add_noise}. $\zeta$ is a scale parameter to control the ratio of signal and noise.

After sampling $y_t$, FDP generates noised heatmaps $y^{hm}_t$ and feature masks $y_t^{mask}$, which can be formulated as:
\begin{equation}
\label{eq:hm_and_mask}
\begin{aligned}
     y_t^{hm}  &= \phi(y_t, \sigma),\\
    y_t^{mask} = \{M^{kps}, & M^{ske}\} = \varphi(y_t, \delta_{kps}, \delta_{ske}), \\
\end{aligned}
\end{equation}
where $\phi(\cdot)$ represents the operation of generating 2D Gaussian heatmaps with center points $y_t$ and sigma parameter $\sigma$ following \cite{xiao2018simple}. $\varphi(\cdot)$ represents the operation of generating 2D keypoints masks $M^{kps}$ and skeleton masks $M^{ske}$ according to keypoints $y_t$ and its corresponding skeleton. $\delta_{kps}$ and $\delta_{ske}$ control the widths of keypoints masks and skeleton masks.

\subsubsection{Model Forward Process}
After diffusing keypoints $y_0$ to noised keypoints $y_t$, diffusion model $f_\theta$ learns to recover keypoints heatmaps from noised heatmaps.
$f_\theta$ includes encoder $E(\cdot)$, spatial-channel cross-attention (SC-CA) module $A(\cdot)$, and decoder $D(\cdot)$.

The Model Forward Process (MFP) is shown in Figure \ref{fig:framework} (Black flow process) and Algorithm \ref{alg:train}.
Given an image $x$ of size ${H}_x\times {W}_x$, noised heatmaps $y^{hm}_t$, feature masks $M^{kps}$ and skeleton masks $M^{ske}$, DiffusionPose first extract deep feature by the encoder as
$x^{fea} = E(x)$, where $E(\cdot)$ represents the encoder of diffusion model.
Following previous work~\cite{sun2019deep}, HRNet pre-trained on ImageNet~\cite{deng2009imagenet} dataset is used as the backbone network to extract features since it is the widely-used benchmark.

Then, keypoints features and skeleton features are extracted by the masking operation as follows:
\begin{equation}
\label{eq:mask}
        x^{kps} = x^{fea} \otimes M^{kps},
        x^{ske} = x^{fea} \otimes M^{ske},
\end{equation}
where $\otimes$ is element-wise multiplication. Next, the cross-attention is applied on $x^{kps}$ and $x^{ske}$ to capture structural information, and the output $x^c$ is further used as the conditions of diffusion decoder, which can be formulated as:
\begin{equation}
\label{eq:condition}
    x^c = C(x^{kps} \oplus A(x^{kps}, x^{ske}, x^{ske}), x^{fea}),
\end{equation}
where $C(\cdot)$ and $\oplus$ represent concatenating features and summing, respectively.
$A(\cdot)$ represents the SC-CA module of DiffusionPose. SC-CA module aims to compute the relationship between keypoints features and skeleton features, which includes spatial attention and channel attention.
The SC-CA module can be denoted as:
\begin{equation}
    A(Q, K, V) = A_{c}(Q, K, V) + A_{s}(Q, K, V),
\end{equation}
where $A_c$ and $A_s$ represent channel group attention and spatial window multi-head attention proposed in~\cite{ding2022davit}.
$A_s$ aims to improve the accuracy of localization since the feature conditions $x^c$ contain noises.
$A_c$ aims to enhance the semantic information of conditions.
$Q$, $K$, and $V$ are query, key, and value tokens, respectively.

After obtaining conditions $x^c$, the diffusion decoder $D(\cdot)$ estimates the keypoints heatmaps $\tilde{y}^{hm}_{t-1}$ as:
\begin{equation}
\label{eq:pred_y0}
    \tilde{y}^{hm}_{t-1} = D({y}_t^{hm}, x^c).
\end{equation}

For the diffusion decoder, we reform the diffusion model in content generation tasks and propose a Structure-Guided Diffusion Decoder (SGDD). To achieve better performance, we further propose high-resolution SGDD.

\begin{figure}[t]
  \centering
  \includegraphics[width=\linewidth]{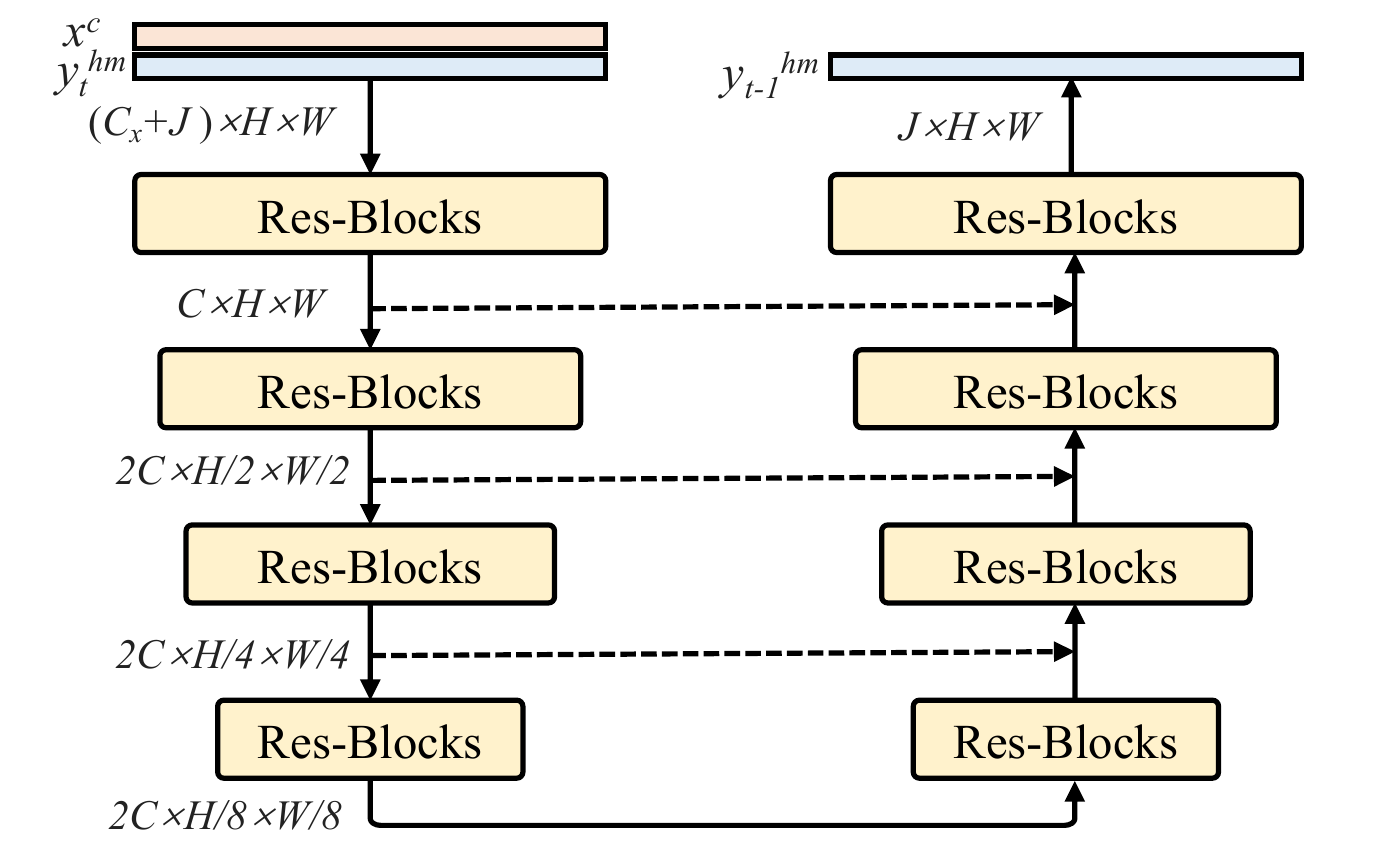}
  \caption{The architecture of Diffusion Decoder in DiffusionPose.}
  \label{fig:unet}
\end{figure}

\textbf{Structure-Guided Diffusion Decoder.} 
The condition $x^c$ with structural information is used to guide the generation of human heatmaps by SGDD. SGDD aims to recover the ground-truth heatmaps from noised heatmaps generated from the forward diffusion process, which contains a neural network with U-Net architecture as the previous works~\cite{ho2020denoising,chen2022generalist}. As shown in Figure~\ref{fig:unet}, given noised heatmap $y_t^{hm}\in \mathbb{R}^{J\times H \times W}$ and condition $x^c \in \mathbb{R}^{C_x\times H \times W}$, SGDD outputs the estimated heatmaps $y_0^{hm} \in \mathbb{R}^{J\times H \times W}$, where $C_x$, $J$, $H\times W$ represent the channel number of $x^c$, joints number, and the size of heatmaps, respectively.

In SGDD, input features and heatmaps are down-sampled to a size of $2C\times \frac{H}{8}\times \frac{W}{8}$ in a progressive way, and up-sampled to the original size of $H\times W$ with the residual connection, where $C\in \{64, 128, 256\}$ is the channel number of features in the decoder. In each scale of SGDD, multiple Res-Blocks~\cite{he2016deep} are stacked to extract features and recover heatmaps. The architecture parameters of SGDD are shown in Table \ref{tab:unet_archi}.

\begin{table}
  \centering
  \renewcommand\tabcolsep{6pt}
  \resizebox{\columnwidth}{!}{
  \begin{tabular}{l|c|c|c|c}
    \toprule
    Stage & 1 & 2 & 3 & 4 \\
    \midrule
    Res-Blocks & 2 & 2 & 2  & 2 \\
    \midrule
    Channels & C & 2C & 2C  & 2C \\
    \midrule
    Feature Size & $H\times W$ & $H/2 \times W/2$ & $H/4 \times W/4$  & $H/8 \times W/8$ \\
    \bottomrule
  \end{tabular}}
    \vspace{0.05cm}
  \caption{The architecture parameters of SGDD. Channel dimension $C\in \{64, 128, 256\}$.}
  \label{tab:unet_archi}
\end{table}

\textbf{High-Resolution SGDD.}
To achieve better performance, SGDD can generate high-resolution heatmaps. As shown in Figure~\ref{fig:unet}, the input size of the diffusion decoder is $H \times W$. In SGDD, $H=H_x/s$ and $W=W_x/s$, where $H_x \times W_x$ is the original size of input image $x$ and $s\in \{1, 2, 4, 8\}$ is the scale of heatmaps. Thus, SGDD can run in choosing different scales.
The diffusion decoding process with higher heatmap resolution could mitigate the quantization error and improve the quality of heatmaps.

\begin{algorithm}[t]
	\renewcommand{\algorithmicrequire}{\textbf{Input:}}
	\renewcommand{\algorithmicensure}{\textbf{Return:}}
	\caption{Training DiffusionPose model $f_\theta$}
	\label{alg:train}
	\begin{algorithmic}[1]
            \REQUIRE image $x$, keypoints labels $y_0$, diffusion step $T$, $f_\theta$ contains encoder $E$ and diffusion decoder $D$.
		\STATE Extract feature $x^{fea}=E(x)$
            \REPEAT
            \STATE $t\sim \text{Uniform}(\{1,...,T\})$
            \STATE $y_t = q(y_t|y_0, \zeta)$ 
            \STATE $y_t^{hm} = \phi(y_t, \sigma)$, $y_0^{hm}=\phi(y_0, \sigma)$ 
            \STATE $y_t^{mask}= \{M^{kps}, M^{ske}\} = \varphi(y_t, \delta_{kps}, \delta_{ske})$ 
            \STATE $x^{kps} = x^{fea}\otimes M^{kps}$, $x^{ske} = x^{fea}\otimes M^{ske}$ 
            \STATE $x^c = C(x^{fea}, x^{kps}\oplus A(x^{kps}, x^{ske}, x^{ske}))$
            \STATE $L_1 = ||D(y_t^{hm}, x^c) - y_0^{hm}||_2$
            \STATE $L_2 = ||conv(x^{fea}) - y_0^{hm}||_2$
            \STATE Take gradient descent step on $\nabla_\theta (L_1 + L_2) $
            \UNTIL converged
	\end{algorithmic}  
\end{algorithm}

\begin{algorithm}[t]
	\renewcommand{\algorithmicrequire}{\textbf{Input:}}
	\renewcommand{\algorithmicensure}{\textbf{Return:}}
	\caption{Inference in $T$ iterative steps}
	\label{alg:inferece}
	\begin{algorithmic}[1]
            \REQUIRE image $x$, diffusion step $T$.
		\STATE Extract feature $x^{fea}=E(x)$
            \STATE $\tilde{y}^{hm}_T = conv(x^{fea})$
            \FOR{$t=T, ... , 1 $}
                \STATE Deocde coordinates $\tilde{y}_t = argmax(\tilde{y}_t^{hm})$
                \STATE $y^{mask}_t= \{M^{kps}, M^{ske}\} = \varphi(\tilde{y}_t, \delta_{kps}, \delta_{ske})$
                \STATE $x^{kps} = x^{fea}\otimes M^{kps}$, $x^{ske} = x^{fea}\otimes M^{ske}$
                \STATE $x^c = C(x^{fea}, x^{kps}\oplus A(x^{kps}, x^{ske}, x^{ske}))$
                \STATE $\tilde{y}_{t-1}^{hm} = D(\tilde{y}^{hm}_t, x^c)$
            \ENDFOR
            \STATE Deocde coordinates $\tilde{y}_0 = argmax(\tilde{y}_0^{hm})$
            \ENSURE $\tilde{y}_0$
            
	\end{algorithmic}  
\end{algorithm}

\subsubsection{Reverse Diffusion Process}
During inference, starting from initialized heatmaps, DiffusionPose estimates the keypoints heatmaps in a progressive way. DiffusionPose adopts the strategy of skipping-step sampling as DDIM~\cite{song2020denoising}, which does not require inference steps equal to the training steps.
The Reverse Diffusion Process (RDP) is shown in Figure \ref{fig:framework} (red flow process) and Algorithm \ref{alg:inferece}, which includes sampling initialization and reverse diffusion inference.

\textbf{Sampling Initialization.}
A good sampling initialization could reduce the inference steps. In DiffusionPose, given the time step $T$, the noised heatmaps $y_T$ are initialized as:
\begin{equation}
    \tilde{y}_T^{hm} = conv(x^{fea}).
\end{equation}
$\tilde{y}_T^{hm}$ can provide better initialization since it is trained with the supervision of ground-truth heatmaps.

\textbf{Reverse Diffusion Inference.}
After initializing noised heatmaps, DiffusionPose estimates ground-truth heatmaps by SGDD. The process is shown in Algorithm \ref{alg:inferece}. Feature masks $y^{mask}$ are generated from $\tilde{y}_T^{hm}$, which is further used to generate conditions $x^c$ as Equation \ref{eq:mask} and \ref{eq:condition}.
With $\tilde{y}_{t}^{hm}$ and $x^c$ as inputs, SGDD estimates the $\tilde{y}_{t-1}^{hm}$ as Equation \ref{eq:pred_y0}.
Then, DiffusionPose iteratively estimates $\tilde{y}_0^{hm}$ by repeating the above steps.
Finally, the keypoints $\tilde{y}_0$ are decoded from $\tilde{y}_0^{hm}$ by $argmax$ function.

\subsection{Loss Function}
DiffusionPose iteratively generates heatmaps from noised heatmaps. Following the previous works~\cite{xiao2018simple,sun2019deep,xu2022vitpose}, $L_2$ loss is used as the loss function. To speed up the inference process, DiffusionPose learns coarse heatmaps from $x^{fea}$ by a convolutional layer $conv$, which is used to initialize heatmaps for inference.
Therefore, the total loss is
\begin{equation}
    L = ||D(y_t^{hm},x^c) - y^{hm}_0||_2 + ||conv(x^{fea}) - y^{hm}_0||_2.
\end{equation}

\begin{table*}
  \centering
  \renewcommand\tabcolsep{9pt}
  \resizebox{\textwidth}{!}{
  \begin{tabular}{c|l|c|cccc|cc}
    \toprule
    \# &Method & Backbone & Params (M) & FLOPs (G) & Resolution & Feature Scale & AP & AR\\
    \midrule
    \multirow{5}{*}{\rotatebox{90}{Benchmarks}}&SimpleBaseline~\cite{xiao2018simple} & ResNet-152 & 68 &  29 & 256x192 & 1/32 & 73.5 & 79.0 \\

    &HRNet~\cite{sun2019deep} & HRNet-W32 & 29 & 8 & 256x192 & 1/4 & 74.4 & 78.9\\
    &HRNet~\cite{sun2019deep} & HRNet-W32 & 29 & 18 & 384x288 & 1/4 & 75.8 & 81.0 \\
    &HRNet~\cite{sun2019deep} & HRNet-W48 & 64 & 16 & 256x192 & 1/4 & 75.1 & 80.4 \\
    &HRNet~\cite{sun2019deep} & HRNet-W48 & 64 & 36 & 384x288 & 1/4 & 76.3 & 81.2 \\
    \midrule
   \multirow{5}{*}{\rotatebox{90}{Refinement}} 
    &DARK~\cite{zhang2020distribution} & HRNet-W32 & 29 & 18 & 384x288 & 1/4 & 76.6 & 81.5 \\
    &DARK~\cite{zhang2020distribution} & HRNet-W48 & 64 & 36 & 384x288 & 1/4 & 76.8 & 81.7 \\
    &UDP~\cite{huang2020devil} + DARK~\cite{zhang2020distribution} & HRNet-W32 & 29 & 16 & 384x288 & 1/4 & 76.8 & 81.5 \\
    &UDP~\cite{huang2020devil} + DARK~\cite{zhang2020distribution} & HRNet-W48 & 64 & 36 & 384x288 & 1/4 & 77.2 & 82.0 \\
    &PoseFix~\cite{moon2019posefix} & W48+Res-152 & 132 & 65 & 384x288 & 1/4 & 77.2 & 82.0 \\
    \midrule
    \multirow{5}{*}{\rotatebox{90}{Transformer}}&TokenPose-L/D24$^\dag$~\cite{li2021tokenpose} & HRNet-W48 & 28 & 11 & 256x192 & 1/4 & 75.8 & 80.9 \\
    &TransPose-H/A6$^\dag$~\cite{yang2021transpose} & HRNet-W48 & 18 & 22 & 256x192 & 1/4 & 75.8 & 80.8 \\
    &HRFormer-B$^\dag$~\cite{yuan2021hrformer} & HRFormer-B & 43 & 14 & 256x192 & 1/4 & 75.6 & 80.8 \\
    &HRFormer-B$^\dag$~\cite{yuan2021hrformer} & HRFormer-B & 43 &  29& 384x288 & 1/4 & 77.2 & 82.0 \\

    &ViTPose-B(+UDP)~\cite{xu2022vitpose} & ViT-B & 86 & 22 & 256x192 & 1/16 & 75.8 & 81.1 \\

    \midrule

    \multirow{4}{*}{\rotatebox{90}{Diffusion}}&\textbf{DiffusionPose}  & HRNet-W32 & 38 & 14 & 256x192  & 1/4 & 75.9 ($\uparrow$1.5) & 81.1\\
    &\textbf{DiffusionPose}  & HRNet-W32 & 38 & 31 & 384x288  & 1/4 & 77.1 ($\uparrow$1.3) & 81.8\\
    &\textbf{DiffusionPose}  & HRNet-W48 & 74 & 22 & 256x192  & 1/4 & 76.7 ($\uparrow$1.6) & 81.7\\
    &\textbf{DiffusionPose}  & HRNet-W48 & 74 & 49 & 384x288  & 1/4 & \textbf{77.6} ($\uparrow$1.3) & \textbf{82.4} \\

    \bottomrule
  \end{tabular}
  }
  \vspace{0.05cm}
  \caption{The comparison of DiffusionPose and SOTA methods on COCO~\cite{lin2014microsoft} dataset. $\dag$ indicates the Transformer-based methods. Other results are cited from \cite{xu2022vitpose}. All results are classified into four categories: Benchmarks, Refinement-based methods, Transformer-based methods, and DiffusionPose. $\uparrow$ shows the improvements compared with benchmarks.}
  \label{tab:coco}
\end{table*}

\section{Experiments}

\subsection{Implementation Details}
The backbone network of DiffusionPose is HRNet~\cite{sun2019deep} with pre-trained weights on ImageNet~\cite{deng2009imagenet}.
All ablation studies are based on HRNet-W32.
During training, Adam optimizer with an initial learning rate of 5e-4 is used.
DiffusionPose is trained by 210 epochs and the learning rate is divided by 10 at the 170th and 210th epochs.
Data augmentations include random crop and random horizontal flip.
DiffusionPose is trained on 8 Tesla A100 GPUs with a batch size of $32\times 8$. During inference, the flip test is used following \cite{sun2019deep,xu2022vitpose}.
Except specifically noted, the input size is $256\times 192$.

\subsection{Datasets and Evaluation Metric}
We evaluate DiffusionPose on COCO~\cite{lin2014microsoft}, CrowdPose~\cite{li2019crowdpose}, and AI Challenge~\cite{wu2019large} datasets, respectively. All ablation studies are finished on the COCO dataset.

\textbf{Datasets.}
COCO~\cite{lin2014microsoft} dataset contains about 200k images and 250k person instances, in which each person is annotated with 17 keypoints. The training and validation subsets are used.
For a fair comparison with \cite{sun2019deep,xu2022vitpose}, the human bounding boxes from a detector with 56.4 mAP are used.
CrowdPose~\cite{li2019crowdpose} contains 20k images and 80k person instances, in which each person is annotated with 14 keypoints. Since more persons with serious occlusions are in an image, CrowdPose is more difficult. For a fair comparison with \cite{yuan2021hrformer,ding20222r}, the ground-truth human bounding boxes when evaluating pose estimation models.
AI Challenge~\cite{wu2019large} contains 350k images and 700k human instances, in which each person is annotated with 14 keypoints. For a fair comparison with \cite{sun2019deep,yuan2021hrformer}, the ground-truth human bounding boxes are used when testing pose estimation models.

\textbf{Evaluation Metrics.}
Following the standard evaluation metrics of COCO datatset~\cite{lin2014microsoft}, the Object Keypoint Similarity (OKS) based metrics are used. In each dataset, AP, AP$_{50}$, AP$_{75}$, AR, AR$_{50}$, and AR$_{75}$ are used. In the CrowdPose dataset, the APs on different crowd-index (easy, medium, hard) are also reported: AP$_{E}$, AP$_{M}$, and AP$_H$.

\begin{table*}
  \centering
  \renewcommand\tabcolsep{12pt}
  \resizebox{\textwidth}{!}{
  \begin{tabular}{l|c|ccccccccc}
    \toprule
    Method & Backbone & AP & AP$_{50}$ & AP$_{75}$ & AR & AR$_{50}$ & AR$_{75}$ & AP$_{E}$ & AP$_{M}$ & AP$_{H}$\\
    \midrule
    OPEC-Net~\cite{qiu2020peeking} & ResNet-101 & 70.6 & 86.8 & 75.6 & - &- &- &- &- &-\\
    HRNet~\cite{sun2019deep} & HRNet-W32 & 71.3 & 91.1 & 77.5 & - & - & - & 80.5 & 71.4 & 62.5\\
    TransPose-H$^\dag$ ~\cite{yang2021transpose} & HRNet-W48 & 71.8 & 91.5 & 77.8 & 75.2 & 92.7 & 80.4 & 79.5 & 72.9 & 62.2 \\
    HRFormer-B$^\dag$ ~\cite{yuan2021hrformer} & HRFormer-B & 72.4 & 91.5 & 77.9 & 75.6 & 92.7 & 81.0 & 80.0 & 73.5 & 62.4 \\
    I$^2$R-Net~\cite{ding20222r} & HRNet-W48 & 72.3 & 92.4 & 77.9 & 76.5 & 93.2 & 81.9 & 79.9 & 73.2 &62.8\\
    I$^2$R-Net$^\dag$ ~\cite{ding20222r} & TransPose-H & 76.3 & 93.5 & 82.2 & 79.1 & 94.0 & 84.4 & 83.2 & 77.0 &67.4\\
    \midrule

    \textbf{DiffusionPose}  & HRNet-W32 & 76.7 & 93.5 & 82.4  & 79.5 & 94.2 & 84.9 & 83.7 & 77.6 & 67.7\\
    \textbf{DiffusionPose}  & HRNet-W48 & \textbf{77.5} & \textbf{93.6} & \textbf{83.5}  & \textbf{80.3} & \textbf{94.5} & \textbf{85.6} & \textbf{84.4} & \textbf{78.5} & \textbf{68.7} \\

    \bottomrule
  \end{tabular}
  }
  \vspace{0.05cm}
  \caption{The comparison of DiffusionPose and SOTA methods on CrowdPose~\cite{li2019crowdpose} dataset. Other results are cited from \cite{ding20222r}.}
  \label{tab:crowdpose}
\end{table*}

\subsection{Comparisons with SOTA Methods}

\textbf{COCO dataset.}
We evaluate DiffusionPose on COCO dataset~\cite{lin2014microsoft} and compare it with SOTA methods.
As shown in Table \ref{tab:coco}, other methods are classified into four categories: Benchmarks, Refinement-based, and Transformer-based methods. Compared with HRNet benchmarks~\cite{sun2019deep}, HRNet-W32-based DiffusionPose achieves improvements of 1.5 and 1.3 AP with input sizes of $256\times 192$ and $384\times 288$, respectively.
Compared with HRNet-W48, DiffusionPose also achieves improvements of 1.6 and 1.3 AP, respectively.
The FLOPs of DiffusionPose are comparable with benchmarks.
Compared with the refinement-based methods, which refine the keypoints heatmaps by learning keypoints offset~\cite{huang2020devil}, adjusting heatmap peaks~\cite{zhang2020distribution}, and extra models refining~\cite{moon2019posefix}, DiffusionPose has the improvements of 0.4 to 1 AP. 
Compared with Transformer-based methods~\cite{li2021tokenpose,yang2021transpose,yuan2021hrformer,xu2022vitpose}, DiffusionPose shows improvements of 0.4 to 1.1 AP with the CNN backbone network.

\textbf{CrowdPose dataset.}
To evaluate DiffusionPose on crowd scenarios, 
we test it on CrowdPose~\cite{li2019crowdpose} dataset and compare it with SOTA methods. The images in the CrowdPose dataset contain multiple persons, which are more difficult to tackle since heavy occlusions. 
As shown in Table~\ref{tab:crowdpose}, DiffusionPose achieves new SOTA results of 76.7 and 77.5 AP with HRNet-W32 and HRNet-W48 as backbone networks, respectively.
Compared with previous SOTA I$^2$R-Net \cite{ding20222r} with TransPose~\cite{yang2021transpose} as backbone network, DiffusionPose has an improvement of 0.8 AP with a smaller HRNet-W48.
Specifically, DiffusionPose achieves improvements of 1.2, 1.5, and 1.3 AP on AP$_E$, AP$_M$, and AP$_H$, respectively.
These results show the stronger ability of DiffusionPose on handling occlusion cases in the crowd scenarios from the scheme of generating heatmaps.

\begin{table}
  \centering
  \renewcommand\tabcolsep{5pt}
  \resizebox{\columnwidth}{!}{
  \begin{tabular}{l|c|cccc}
    \toprule
    Method & Backbone & AP & AP$_{50}$ & AR & AR$_{50}$ \\
    \midrule
    SimpleBaseline~\cite{xiao2018simple} & ResNet-50 & 28.0 & 71.6 & 32.1& 74.1 \\
    SimpleBaseline~\cite{xiao2018simple} & ResNet-101& 29.4 & 73.6 & 33.7 & 76.3 \\
    SimpleBaseline~\cite{xiao2018simple} & ResNet-152& 29.9 & 73.8 & 34.3 & 76.9 \\

    HRNet~\cite{sun2019deep} & HRNet-W32 & 32.3 & 76.2 & 36.6 & 78.9 \\
    HRNet~\cite{sun2019deep} & HRNet-W48 & 33.5 & 78.0 & 37.9 & 80.0 \\

    HRFormer$^\dag$~\cite{yuan2021hrformer} & HRFomer-S & 31.6 & 75.9 & 35.8 & 78.0 \\
    HRFormer$^\dag$~\cite{yuan2021hrformer} & HRFomer-B & 34.4 & 78.3 & 38.7 & 80.9 \\


    \midrule
    \textbf{DiffusionPose} & HRNet-W32 & 34.7 & 78.0 & 39.4 & 80.1 \\
    \textbf{DiffusionPose} & HRNet-W48 & \textbf{35.6} & \textbf{79.1} & \textbf{40.1} & \textbf{81.1}\\
    \bottomrule
  \end{tabular}}
  
  \vspace{0.05cm}
  \caption{Comparison with SOTA methods on AI Challenger~\cite{wu2019large}.}
  \label{tab:aic}
\end{table}

\textbf{AI Challenge dataset.}
We evaluate DiffusionPose on the challenging in-the-wild dataset: AI Challenge~\cite{wu2019large}.
As shown in Tabel~\ref{tab:aic}, DiffusionPose achieves state-of-the-art performances by 34.7 and 35.6 AP with HRNet-W32 and HRNet-W48 as the backbone networks, respectively. Compared with benchmark HRNet~\cite{sun2019deep}, DiffusionPose brings improvements of 2.4 and 2.1 AP on W32 and W48, respectively. Compared with HRFormer~\cite{yuan2021hrformer}, DiffusionPose outperforms it by 1.2 AP.
These results show the effectiveness and generalization ability of DiffusionPose.

\subsection{Ablation Study}

\begin{table}
  \centering
  \renewcommand\tabcolsep{2pt}
  \resizebox{\columnwidth}{!}{
  \begin{tabular}{l|c|c|cc}
    \toprule
    Model & Diffusion & Conditions $x^c$ & AP & AR\\
    \midrule
    Baseline & $\times$  & $-$ & 74.4 & 78.9\\
    DiffusionPose & $\checkmark$ & $x^{fea}$  & 75.0 & 80.0\\
    DiffusionPose & $\checkmark$ & $C(x^{fea},x^{kps})$ & 75.2 & 80.3\\
    DiffusionPose & $\checkmark$  & $C(x^{fea},A(x^{kps}, x^{ske}, x^{ske}))$ & \textbf{75.4} & \textbf{80.5}\\

    \bottomrule
  \end{tabular}}
    \vspace{0.05cm}
  \caption{The ablation study of main components of SGDD. $C(\cdot)$ denotes concatenating operation and $A$ is the SC-CA module.}
  \label{tab:ab_module}
\end{table}

\begin{figure}[t]
  \centering
  \includegraphics[width=\linewidth]{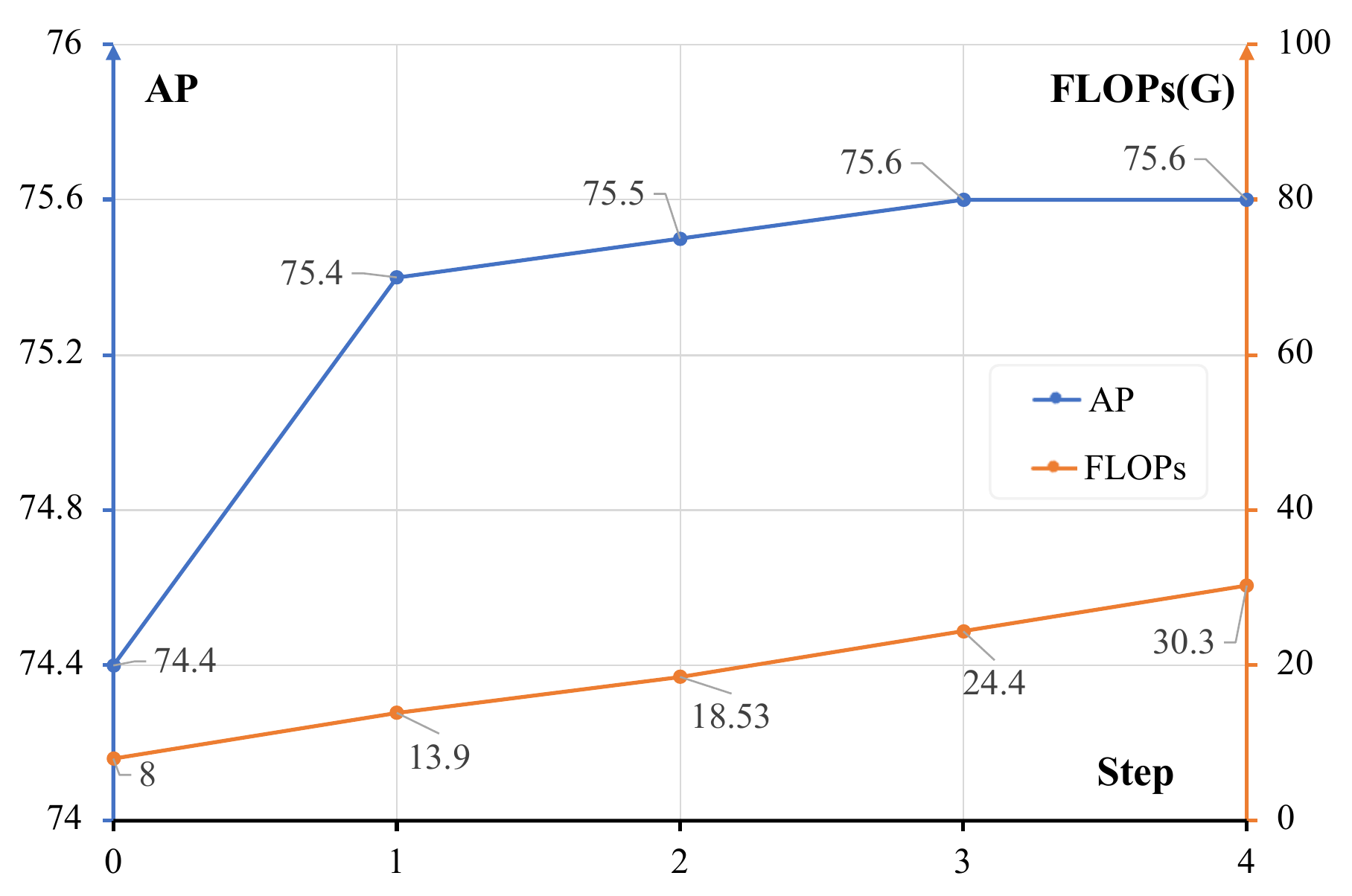}
  \caption{The performances and FLOPs of DiffusionPose with the increasing of sampling steps. [Best viewed in color]}
  \label{fig:diff_step}
\end{figure}

\textbf{Main Components of SGDD.}
The ablation study of the main components of the SGDD is shown in Table \ref{tab:ab_module}. In Table \ref{tab:ab_module}, the default diffusion inference step is 1.
The baseline of HRNet-W32 achieves 74.4 AP and 78.9 AR on COCO~\cite{lin2014microsoft} dataset. DiffusionPose with image features as conditions achieves 75.0 AP. Adding keypoints features as conditions results in an improvement of 0.2 AP. The final version of DiffusionPose achieves 75.4 AP with improvements of 1 AP, with the SC-CA module to capture structural information from kepoints and skeleton features.

\textbf{Diffusion Steps and Computation Costs.}
DiffusionPose iteratively generates high-quality keypoints heatmaps. Figure \ref{fig:diff_step} shows the increasing performances and FLOPs of DiffusionPose with increasing inference steps.
During inference, Diffusion initializes input heatmaps from coarse heatmaps by $conv(x^{fea})$, which is supervised ground-truth heatmaps and provides a good starting point for the diffusion model. As shown in Figure \ref{fig:diff_step}, just one reverse step achieves 75.4 AP. With more inference steps, the gains are about 0.1-0.2 AP and the FLOPs increase linearly. Thus, we suggest 1 step during inference for the trade-off of performance and cost.
Compared with previous image/video generation models~\cite{ho2020denoising,holmquist2022diffpose} and detection/segmentation models~\cite{li2022srdiff,gu2022diffusioninst}, DiffusionPose achieves better results with fewer inference steps from good initialized heatmaps $y^{hm}_T$.

\textbf{High-resolution Diffusion Decoder.}
DiffusionPose supports generating heatmaps in different resolutions.
Usually, high-resolution heatmaps result in better performance in 2D human pose estimation as \cite{sun2019deep,yuan2021hrformer,huang2020devil}.
We also conduct the ablation study of different resolutions of generated heatmaps. As shown in Table \ref{tab:ab_resolution}, DiffusionPose achieves better performances with a higher resolution. Considering the trade-off between computation costs and performance, we suggest adopting the setting of $s=2$, which achieves 75.9 AP with the input image size of $256\times 192$.

\begin{table}
  \centering
  \renewcommand\tabcolsep{12pt}
  \resizebox{\columnwidth}{!}{
  \begin{tabular}{l|cc|cc}
    \toprule
    Model  &  $s$ & Resolution & AP & AR\\
    \midrule
    \multirow{4}{*}{DiffusionPose}  & 8 & 32x24 & 72.2  & 78.1\\
     & 4 & 64x48  & 75.4 & 80.5\\
     & 2 & 128x96 & 75.9 & 81.1\\
     & 1 & 256x192 & 76.1 & 81.3\\
    \bottomrule
  \end{tabular}}
    \vspace{0.05cm}
  \caption{The ablation study of SGDD on different resolutions.}
  \label{tab:ab_resolution}
\end{table}

\begin{table}
  \centering
  \renewcommand\tabcolsep{8pt}
  \resizebox{\columnwidth}{!}{
  \begin{tabular}{l|c|c|cc}
    \toprule
    Model & Backbone & Signal Scale $\zeta$& AP & AR\\
    \midrule
    \multirow{5}{*}{DiffusionPose} & \multirow{5}{*}{HRNet-W32} & 0.01 &  75.3 & 80.5\\
    & & 0.05  & \textbf{75.4} & \textbf{80.5}\\
     & & 0.10  & 75.2 & 80.4\\
     & & 0.50 & 71.6 & 78.0\\
    & & 1.00 & 70.7 & 77.5\\

    \bottomrule
  \end{tabular}}
    \vspace{0.05cm}
  \caption{The ablation study of diffusion signal scale $\zeta$.}
  \label{tab:ab_scale}
\end{table}

\textbf{Signal Scale.}
In the diffusion process, hyper-parameter $\zeta$ controls the signal scale, which greatly affects the performance of DiffusionPose. As shown in Table \ref{tab:ab_scale}, DiffusionPose achieves the best performance when $\zeta = 0.05$. Similar to segmentation task~\cite{chen2022generalist}, a smaller scale factor of 0.05 is better than the standard scaling of 1.0 in previous work~\cite{chen2022analog}.

\begin{figure}[t]
  \centering
  \includegraphics[width=\linewidth]{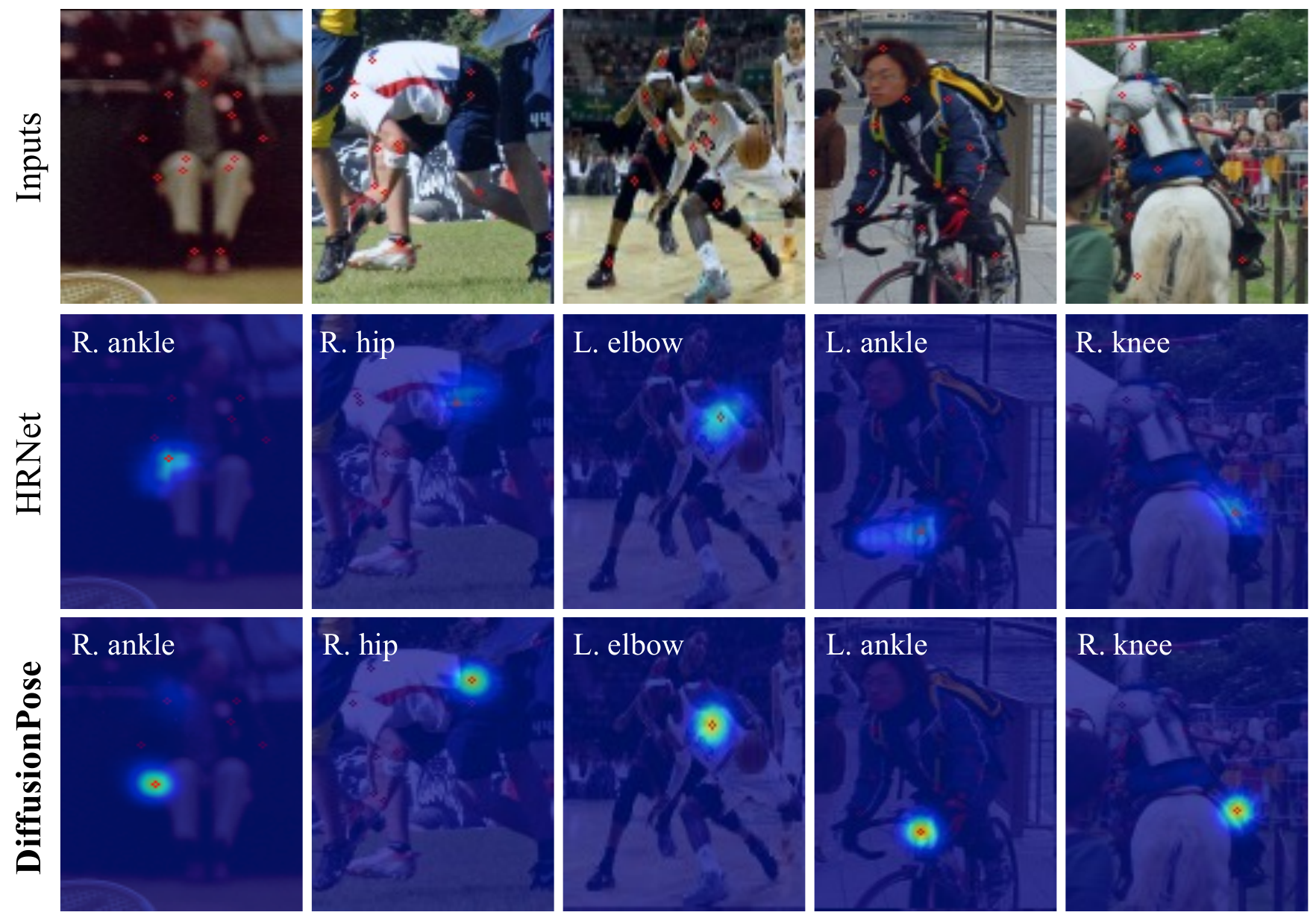}
  \caption{The joints heatmaps generated by traditional scheme~\cite{sun2019deep} and our diffusion-generation scheme on the challenging cases.}
  \label{fig:vis}
\end{figure}

\textbf{Visualization Analysis.}
We compare the generated keypoints heatmaps by DiffusionPose and benchmark HRNet~\cite{sun2019deep} in Figure \ref{fig:vis}. DiffusionPose with HRNet as backbone network achieves better pose results on these challenging cases from CrowdPose~\cite{li2019crowdpose} dataset by generating better keypoints heatmaps.

\subsection{Limitations and Discussion}
Although DiffusionPose successfully introduces the diffusion models into 2D human pose estimation and achieves great results on three datasets, there are some limitations. 1) As shown in Figure~\ref{fig:diff_step}, the FLOPs of DiffusionPose increase linearly with the inference steps, which limits the application of the diffusion model. We introduce an insight that sampling from a good initialization could alleviate this problem. How to generate content in one step is still worth exploring. 2) DiffusionPose extracts deep features by an image encoder as the conditions of the diffusion model. Directly using original images as the conditions is worth exploring, which could reduce the computation costs.

\section{Conclusion}
In this paper, we formulate 2D human pose estimation as a new scheme that generates heatmaps from noised heatmaps. We propose DiffusionPose, the first work to apply the diffusion model to 2D human pose estimation. To generate high-quality heatmaps, DiffusionPose diffuses keypoints to random distribution by adding noises and estimates 2D heatmaps from noised heatmaps in a progressive denoising way.
Furthermore, we propose the structure-guided diffusion decoder (SGDD) and high-resolution SGDD for DiffusionPose, leading it to achieve significant improvements on three widely-used datasets.

\clearpage

{\small
\bibliographystyle{ieee_fullname}
\bibliography{ref}

\begin{thebibliography}{10}\itemsep=-1pt

\bibitem{baranchuk2021label}
Dmitry Baranchuk, Ivan Rubachev, Andrey Voynov, Valentin Khrulkov, and Artem
  Babenko.
\newblock Label-efficient semantic segmentation with diffusion models.
\newblock {\em ICLR}, 2022.

\bibitem{brempong2022denoising}
Emmanuel~Asiedu Brempong, Simon Kornblith, Ting Chen, Niki Parmar, Matthias
  Minderer, and Mohammad Norouzi.
\newblock Denoising pretraining for semantic segmentation.
\newblock In {\em CVPR}, pages 4175--4186, 2022.

\bibitem{cao2017realtime}
Zhe Cao, Tomas Simon, Shih-En Wei, and Yaser Sheikh.
\newblock Realtime multi-person 2d pose estimation using part affinity fields.
\newblock In {\em CVPR}, 2017.

\bibitem{chen2022diffusiondet}
Shoufa Chen, Peize Sun, Yibing Song, and Ping Luo.
\newblock Diffusiondet: Diffusion model for object detection.
\newblock {\em arXiv preprint arXiv:2211.09788}, 2022.

\bibitem{chen2022generalist}
Ting Chen, Lala Li, Saurabh Saxena, Geoffrey Hinton, and David~J Fleet.
\newblock A generalist framework for panoptic segmentation of images and
  videos.
\newblock {\em arXiv preprint arXiv:2210.06366}, 2022.

\bibitem{chen2022analog}
Ting Chen, Ruixiang Zhang, and Geoffrey Hinton.
\newblock Analog bits: Generating discrete data using diffusion models with
  self-conditioning.
\newblock {\em arXiv preprint arXiv:2208.04202}, 2022.

\bibitem{cheng2020higherhrnet}
Bowen Cheng, Bin Xiao, Jingdong Wang, Honghui Shi, Thomas~S Huang, and Lei
  Zhang.
\newblock Higherhrnet: Scale-aware representation learning for bottom-up human
  pose estimation.
\newblock In {\em CVPR}, pages 5386--5395, 2020.

\bibitem{deng2009imagenet}
Jia Deng, Wei Dong, Richard Socher, Li-Jia Li, Kai Li, and Li Fei-Fei.
\newblock Imagenet: A large-scale hierarchical image database.
\newblock In {\em CVPR}, pages 248--255. Ieee, 2009.

\bibitem{ding2022davit}
Mingyu Ding, Bin Xiao, Noel Codella, Ping Luo, Jingdong Wang, and Lu Yuan.
\newblock Davit: Dual attention vision transformers.
\newblock In {\em ECCV}, pages 74--92. Springer, 2022.

\bibitem{ding20222r}
Yiwei Ding, Wenjin Deng, Yinglin Zheng, Pengfei Liu, Meihong Wang, Xuan Cheng,
  Jianmin Bao, Dong Chen, and Ming Zeng.
\newblock I$^{2}${R}-{N}et: Intra-and inter-human relation network for
  multi-person pose estimation.
\newblock {\em IJCAI}, 2022.

\bibitem{fieraru2018learning}
Mihai Fieraru, Anna Khoreva, Leonid Pishchulin, and Bernt Schiele.
\newblock Learning to refine human pose estimation.
\newblock In {\em CVPRW}, pages 205--214, 2018.

\bibitem{geng2021bottom}
Zigang Geng, Ke Sun, Bin Xiao, Zhaoxiang Zhang, and Jingdong Wang.
\newblock Bottom-up human pose estimation via disentangled keypoint regression.
\newblock In {\em CVPR}, pages 14676--14686, 2021.

\bibitem{gong2022diffpose}
Jia Gong, Lin~Geng Foo, Zhipeng Fan, Qiuhong Ke, Hossein Rahmani, and Jun Liu.
\newblock Diffpose: Toward more reliable 3d pose estimation.
\newblock {\em arXiv preprint arXiv:2211.16940}, 2022.

\bibitem{gu2022diffusioninst}
Zhangxuan Gu, Haoxing Chen, Zhuoer Xu, Jun Lan, Changhua Meng, and Weiqiang
  Wang.
\newblock Diffusioninst: Diffusion model for instance segmentation.
\newblock {\em arXiv preprint arXiv:2212.02773}, 2022.

\bibitem{he2016deep}
Kaiming He, Xiangyu Zhang, Shaoqing Ren, and Jian Sun.
\newblock Deep residual learning for image recognition.
\newblock In {\em CVPR}, pages 770--778, 2016.

\bibitem{ho2020denoising}
Jonathan Ho, Ajay Jain, and Pieter Abbeel.
\newblock Denoising diffusion probabilistic models.
\newblock In {\em NeurIPS}, pages 6840--6851, 2020.

\bibitem{ho2022video}
Jonathan Ho, Tim Salimans, Alexey Gritsenko, William Chan, Mohammad Norouzi,
  and David~J Fleet.
\newblock Video diffusion models.
\newblock {\em NeurIPS}, 2022.

\bibitem{holmquist2022diffpose}
Karl Holmquist and Bastian Wandt.
\newblock Diffpose: Multi-hypothesis human pose estimation using diffusion
  models.
\newblock {\em arXiv preprint arXiv:2211.16487}, 2022.

\bibitem{huang2020devil}
Junjie Huang, Zheng Zhu, Feng Guo, and Guan Huang.
\newblock The devil is in the details: Delving into unbiased data processing
  for human pose estimation.
\newblock In {\em CVPR}, pages 5700--5709, 2020.

\bibitem{kreiss2019pifpaf}
Sven Kreiss, Lorenzo Bertoni, and Alexandre Alahi.
\newblock Pifpaf: Composite fields for human pose estimation.
\newblock In {\em CVPR}, pages 11977--11986, 2019.

\bibitem{li2022srdiff}
Haoying Li, Yifan Yang, Meng Chang, Shiqi Chen, Huajun Feng, Zhihai Xu, Qi Li,
  and Yueting Chen.
\newblock Srdiff: Single image super-resolution with diffusion probabilistic
  models.
\newblock {\em Neurocomputing}, 479:47--59, 2022.

\bibitem{li2021human}
Jiefeng Li, Siyuan Bian, Ailing Zeng, Can Wang, Bo Pang, Wentao Liu, and Cewu
  Lu.
\newblock Human pose regression with residual log-likelihood estimation.
\newblock In {\em ICCV}, pages 11025--11034, 2021.

\bibitem{li2019crowdpose}
Jiefeng Li, Can Wang, Hao Zhu, Yihuan Mao, Hao-Shu Fang, and Cewu Lu.
\newblock Crowdpose: Efficient crowded scenes pose estimation and a new
  benchmark.
\newblock In {\em CVPR}, pages 10863--10872, 2019.

\bibitem{li2021tokenpose}
Yanjie Li, Shoukui Zhang, Zhicheng Wang, Sen Yang, Wankou Yang, Shu-Tao Xia,
  and Erjin Zhou.
\newblock Tokenpose: Learning keypoint tokens for human pose estimation.
\newblock In {\em ICCV}, pages 11313--11322, 2021.

\bibitem{lin2014microsoft}
Tsung-Yi Lin, Michael Maire, Serge Belongie, James Hays, Pietro Perona, Deva
  Ramanan, Piotr Doll{\'a}r, and C~Lawrence Zitnick.
\newblock Microsoft coco: Common objects in context.
\newblock In {\em ECCV}, pages 740--755. Springer, 2014.

\bibitem{liu2020disentangling}
Ziyu Liu, Hongwen Zhang, Zhenghao Chen, Zhiyong Wang, and Wanli Ouyang.
\newblock Disentangling and unifying graph convolutions for skeleton-based
  action recognition.
\newblock In {\em CVPR}, pages 143--152, 2020.

\bibitem{metzer2022latent}
Gal Metzer, Elad Richardson, Or Patashnik, Raja Giryes, and Daniel Cohen-Or.
\newblock Latent-nerf for shape-guided generation of 3d shapes and textures.
\newblock {\em arXiv preprint arXiv:2211.07600}, 2022.

\bibitem{moon2019posefix}
Gyeongsik Moon, Ju~Yong Chang, and Kyoung~Mu Lee.
\newblock Posefix: Model-agnostic general human pose refinement network.
\newblock In {\em CVPR}, pages 7773--7781, 2019.

\bibitem{newell2016stacked}
Alejandro Newell, Kaiyu Yang, and Jia Deng.
\newblock Stacked hourglass networks for human pose estimation.
\newblock In {\em ECCV}, pages 483--499. Springer, 2016.

\bibitem{nichol2022glide}
Alexander~Quinn Nichol, Prafulla Dhariwal, Aditya Ramesh, Pranav Shyam, Pamela
  Mishkin, Bob Mcgrew, Ilya Sutskever, and Mark Chen.
\newblock Glide: Towards photorealistic image generation and editing with
  text-guided diffusion models.
\newblock In {\em ICML}, pages 16784--16804. PMLR, 2022.

\bibitem{qiu2020peeking}
Lingteng Qiu, Xuanye Zhang, Yanran Li, Guanbin Li, Xiaojun Wu, Zixiang Xiong,
  Xiaoguang Han, and Shuguang Cui.
\newblock Peeking into occluded joints: A novel framework for crowd pose
  estimation.
\newblock In {\em ECCV}, pages 488--504. Springer, 2020.

\bibitem{qiu2019learning}
Zhongwei Qiu, Kai Qiu, Jianlong Fu, and Dongmei Fu.
\newblock Learning recurrent structure-guided attention network for
  multi-person pose estimation.
\newblock In {\em ICME}, pages 418--423. IEEE, 2019.

\bibitem{qiu2020dgcn}
Zhongwei Qiu, Kai Qiu, Jianlong Fu, and Dongmei Fu.
\newblock Dgcn: Dynamic graph convolutional network for efficient multi-person
  pose estimation.
\newblock In {\em AAAI}, volume~34, pages 11924--11931, 2020.

\bibitem{qiu2022learning}
Zhongwei Qiu, Huan Yang, Jianlong Fu, and Dongmei Fu.
\newblock Learning spatiotemporal frequency-transformer for compressed video
  super-resolution.
\newblock In {\em ECCV}, pages 257--273. Springer, 2022.

\bibitem{qiu2023psvt}
Zhongwei Qiu, Qiansheng Yang, Jian Wang, Haocheng Feng, Junyu Han, Errui Ding,
  Chang Xu, Dongmei Fu, and Jingdong Wang.
\newblock Psvt: End-to-end multi-person 3d pose and shape estimation with
  progressive video transformers.
\newblock In {\em CVPR}, pages 21254--21263, 2023.

\bibitem{saharia2022image}
Chitwan Saharia, Jonathan Ho, William Chan, Tim Salimans, David~J Fleet, and
  Mohammad Norouzi.
\newblock Image super-resolution via iterative refinement.
\newblock {\em TPAMI}, 2022.

\bibitem{shi2022end}
Dahu Shi, Xing Wei, Liangqi Li, Ye Ren, and Wenming Tan.
\newblock End-to-end multi-person pose estimation with transformers.
\newblock In {\em CVPR}, pages 11069--11078, 2022.

\bibitem{song2020denoising}
Jiaming Song, Chenlin Meng, and Stefano Ermon.
\newblock Denoising diffusion implicit models.
\newblock In {\em ICLR}, 2021.

\bibitem{song2020score}
Yang Song, Jascha Sohl-Dickstein, Diederik~P Kingma, Abhishek Kumar, Stefano
  Ermon, and Ben Poole.
\newblock Score-based generative modeling through stochastic differential
  equations.
\newblock {\em ICLR}, 2021.

\bibitem{sun2019deep}
Ke Sun, Bin Xiao, Dong Liu, and Jingdong Wang.
\newblock Deep high-resolution representation learning for human pose
  estimation.
\newblock In {\em CVPR}, pages 5693--5703, 2019.

\bibitem{sun2017compositional}
Xiao Sun, Jiaxiang Shang, Shuang Liang, and Yichen Wei.
\newblock Compositional human pose regression.
\newblock In {\em ICCV}, pages 2602--2611, 2017.

\bibitem{wang2022contextual}
Dongkai Wang and Shiliang Zhang.
\newblock Contextual instance decoupling for robust multi-person pose
  estimation.
\newblock In {\em CVPR}, pages 11060--11068, 2022.

\bibitem{wang2020graph}
Jian Wang, Xiang Long, Yuan Gao, Errui Ding, and Shilei Wen.
\newblock Graph-pcnn: Two stage human pose estimation with graph pose
  refinement.
\newblock In {\em ECCV}, pages 492--508. Springer, 2020.

\bibitem{wang2022multi}
Yunke Wang, Bo Du, and Chang Xu.
\newblock Multi-tailed vision transformer for efficient inference.
\newblock {\em arXiv preprint arXiv:2203.01587}, 2022.

\bibitem{wang2023learning}
Yunke Wang, Xiyu Wang, Dung Dinh~Anh, Bo Du, and Chang Xu.
\newblock Learning to schedule in diffusion probablisitic models.
\newblock In {\em KDD}, 2023.

\bibitem{wei2020point}
Fangyun Wei, Xiao Sun, Hongyang Li, Jingdong Wang, and Stephen Lin.
\newblock Point-set anchors for object detection, instance segmentation and
  pose estimation.
\newblock In {\em ECCV}, pages 527--544. Springer, 2020.

\bibitem{wu2019large}
Jiahong Wu, He Zheng, Bo Zhao, Yixin Li, Baoming Yan, Rui Liang, Wenjia Wang,
  Shipei Zhou, Guosen Lin, Yanwei Fu, et~al.
\newblock Large-scale datasets for going deeper in image understanding.
\newblock In {\em ICME}, pages 1480--1485. IEEE, 2019.

\bibitem{xiao2018simple}
Bin Xiao, Haiping Wu, and Yichen Wei.
\newblock Simple baselines for human pose estimation and tracking.
\newblock In {\em ECCV}, pages 466--481, 2018.

\bibitem{xu2021h}
Hongyi Xu, Thiemo Alldieck, and Cristian Sminchisescu.
\newblock H-nerf: Neural radiance fields for rendering and temporal
  reconstruction of humans in motion.
\newblock {\em NeurIPS}, 34:14955--14966, 2021.

\bibitem{xu2022vitpose}
Yufei Xu, Jing Zhang, Qiming Zhang, and Dacheng Tao.
\newblock Vitpose: Simple vision transformer baselines for human pose
  estimation.
\newblock In {\em NeurIPS}, 2022.

\bibitem{yang2023explicit}
Jie Yang, Ailing Zeng, Shilong Liu, Feng Li, Ruimao Zhang, and Lei Zhang.
\newblock Explicit box detection unifies end-to-end multi-person pose
  estimation.
\newblock In {\em ICLR}, 2023.

\bibitem{yang2021transpose}
Sen Yang, Zhibin Quan, Mu Nie, and Wankou Yang.
\newblock Transpose: Keypoint localization via transformer.
\newblock In {\em ICCV}, pages 11802--11812, 2021.

\bibitem{yuan2021hrformer}
Yuhui Yuan, Rao Fu, Lang Huang, Weihong Lin, Chao Zhang, Xilin Chen, and
  Jingdong Wang.
\newblock Hrformer: High-resolution transformer for dense prediction.
\newblock In {\em NeurIPS}, 2021.

\bibitem{zhang2020distribution}
Feng Zhang, Xiatian Zhu, Hanbin Dai, Mao Ye, and Ce Zhu.
\newblock Distribution-aware coordinate representation for human pose
  estimation.
\newblock In {\em CVPR}, pages 7093--7102, 2020.

\bibitem{zheng2019deephuman}
Zerong Zheng, Tao Yu, Yixuan Wei, Qionghai Dai, and Yebin Liu.
\newblock Deephuman: 3d human reconstruction from a single image.
\newblock In {\em ICCV}, pages 7739--7749, 2019.

\bibitem{zhou2019objects}
Xingyi Zhou, Dequan Wang, and Philipp Kr{\"a}henb{\"u}hl.
\newblock Objects as points.
\newblock {\em arXiv preprint arXiv:1904.07850}, 2019.

\end{thebibliography}
}

\end{document}